\pdfoutput=1

\documentclass[11pt]{article}
\usepackage{amsfonts,amssymb}
\usepackage[]{acl}
\usepackage{times}
\usepackage{latexsym}
\usepackage{multirow}
\usepackage{MnSymbol}
\usepackage[T1]{fontenc}
\usepackage{booktabs}
\usepackage{amsmath}

\usepackage[utf8]{inputenc}
\usepackage{subfigure}
\usepackage{parskip}
\usepackage{graphicx}
\usepackage{float}
\usepackage{caption}
\usepackage{microtype}

\title{Holistic Sentence Embeddings for Better Out-of-Distribution Detection }

\author{Sishuo Chen\textsuperscript{1}, Xiaohan Bi\textsuperscript{1}, Rundong Gao\textsuperscript{1}, Xu Sun\textsuperscript{2} \\
  \textsuperscript{1}Center for Data Science, Peking University\\
  \textsuperscript{2}MOE Key Laboratory of Computational Linguistics, School of Computer Science, \\ Peking University\\
    \texttt{\{chensishuo,xusun\}@pku.edu.cn} \quad
  \texttt{\{bxh,gaord20\}@stu.pku.edu.cn} }

\begin{document}
\maketitle

\begin{abstract}

Detecting out-of-distribution (OOD) instances is significant for the safe deployment of NLP models. 
Among recent textual OOD detection works based on  pretrained language models (PLMs), distance-based methods have shown superior performance. 
However, they estimate sample distance scores in the last-layer CLS embedding space and thus do not make full use of linguistic information underlying in PLMs. 
To address the issue, we propose to boost OOD detection by deriving more holistic sentence embeddings. 
On the basis of the observations that token averaging and layer combination contribute to improving OOD detection, we propose a simple embedding approach named \textit{Avg-Avg}, which  averages all token representations from each intermediate layer as the sentence embedding  and significantly surpasses the state-of-the-art on a comprehensive suite of benchmarks by a 9.33\% FAR95 margin. 
Furthermore, our analysis demonstrates that it indeed helps preserve general linguistic knowledge in fine-tuned PLMs and substantially benefits detecting background shifts. The simple yet effective embedding method can be applied to fine-tuned PLMs with negligible extra costs, providing a free gain in OOD detection. 
Our code is available at \url{https://github.com/lancopku/Avg-Avg}.

\end{abstract}

\section{Introduction}

Pretrained language models have achieved remarkable performance on various NLP tasks under the assumption that the train and test samples are drawn from the same distribution \citep{glue}. However, in real-life applications such as dialogue systems and clinical text processing, it is inevitable for models to make predictions on out-of-distribution (OOD) samples, which may result in fatally unreasonable predictions \citep{hendrycks2020pretrained}. Therefore, it is crucial for fine-tuned PLMs to automatically detect OOD inputs.
 
Among recent works on textual OOD detection, distance-based methods have received much attention due to their superior performance \citep{podolskiy2021revisiting,contrastive_nlp_ood}. They calculate the sample distance to the training-data distribution as the uncertainty measure for OOD detection.  In these approaches, the distance scores are usually calculated in the space of the last-layer CLS vectors (i.e., the inputs to the classification head) produced by fine-tuned PLMs. As known, the CLS  embedding space is optimized for the in-distribution classification task during fine-tuning, thus not necessarily optimal for OOD detection. 

In this paper, we investigate how to derive sentence embeddings suitable for OOD detection from fine-tuned PLMs. Motivated by the token averaging and layer combination techniques proposed to enhance unsupervised sentence embeddings \cite{su2021whitening,huang-etal-2021-whiteningbert-easy}, we apply them to OOD detection and make two intriguing empirical findings:   (1) averaging all token representations outperforms the standard practice only using the CLS vector; (2) combining token representations from all intermediate layers brings further improvements. 
These observations lead to an extremely simple yet effective pooling technique: averaging all token representations in each intermediate layer as the sentence embedding for OOD detection.

We name the all-layer-all-token pooling technique \textit{Avg-Avg} and demonstrate that it consistently uplifts the OOD detection performance of BERT \cite{devlin-etal-2019-bert} and RoBERTa \cite{roberta} models on a comprehensive suite of textual OOD detection benchmarks. 
Further investigations into the rationales behind the improvement show that \textit{Avg-Avg} effectively helps reserve general linguistic information in the feature space and benefits detecting background shifts. 
In summary, our proposal serves as a plug-and-play post-processing technique to improve the capability of fine-tuned PLMs to detect OOD instances and reveals that it is a promising direction to boost textual OOD detection via deriving more holistic representations.

\section{\textit{Avg-Avg}: Holistic Sentence Embedding for Better OOD Detection} \label{sec_method}

\subsection{Preliminaries}

Modern pretrained language models have been developed based on the Transformer \cite{DBLP:conf/nips/VaswaniSPUJGKP17} architecture. Given a sentence $S = \{t_1,t_2,\ldots,t_n\}$ as the input, an $L$-layer Transformer-based PLM yields a series of hidden vectors $ \mathbb{H} = \{H_0,H_1,\ldots,H_L\}$, where $H_i = \left[ h_i^1,h_i^2,\ldots,h_i^n\right] (0<i\leq L)$ are the embedding vectors for each token in $S$ in the $i$-th Transformer layer and $H_0$ denotes the static token embeddings.

\subsection{Methodology}

In the pretraining-finetuning paradigm, the CLS token is usually put at the beginning of $S$, and the corresponding vector produced by the last Transformer layer $h_L^1$  is fed into the classification head for fine-tuning. In existing works, the CLS vector $h_L^1$ is regarded as the sentence representation, and OOD detection is conducted in the corresponding embedding space \cite{podolskiy2021revisiting,contrastive_nlp_ood}. 
Such a practice does not fully exploit linguistic information contained in $\mathbb{H}$. 
Consequently, we resort to two pooling strategies to derive more holistic sentence representations:

\begin{itemize}
    \item \textit{Intra-Layer Token Averaging}: For the $i$-th layer, we average hidden vectors for all tokens as the pooled representation $P_i$, i.e., $P_i = \frac{1}{n}\sum_{j=1}^n h_i^j$, to replace the default $P_i = h_i^1$.
    
    \item \textit{Inter-Layer Combination}: For given intermediate pooled representations $P_1,P_2,\ldots,P_L$, we perform layer combination to obtain the final pooled sentence representation $P$ for OOD detection: $P= \frac{1}{|M|} \sum_i P_i, i \in M$, where $M \subseteq \{1,2,\ldots,L\}$ denotes the subset of intermediate layers for combination.
\end{itemize}

In our embedding approach \textit{Avg-Avg}, token averaging is performed for intra-layer pooling; all layers are chosen for layer combination, in other words, $M=\{1,2,\ldots,L\}$. Table \ref{tab:pilot} shows the rationality of our choice: for a RoBERTa-based model fine-tuned on the SST-2 sentiment analysis dataset, \textit{Avg-Avg} significantly
outperforms other pooling strategies for detecting 20 Newsgroup (20NG) samples as OOD data, including the default last-layer CLS pooling and the \textit{first-last-avg} pooling used for unsupervised sentence embedding  \citep{su2021whitening}.

\begin{table}[t]
\centering
\resizebox{0.45\textwidth}{!}{
\begin{tabular}{@{}cccc@{}}
\toprule
\textbf{Intra-Layer} & \textbf{Inter-Layer} & \textbf{AUROC\%} & \textbf{Remark} \\ \midrule
CLS            & L12              &     90.48  & Default         \\
Avg                  & L12                  &  93.17     &   -      \\
CLS            & All Layers           &   94.34   &   -     \\
Avg                  & L1+L12               &    98.65   &   \textit{first-last-avg}      \\
Avg                  & All Layers           &   99.99    &    \textit{Avg-Avg} (Ours)     \\ \bottomrule
\end{tabular}}
\caption{The performance of different pooling strategies on the SST-2 v.s. 20NG benchmark. Mahalanobis distance \cite{maha} is the OOD detection method. Avg denotes token average pooling and L12 denotes the 12th layer (the last)  of the RoBERTa model. These results are exploratory and the superiority of \textit{Avg-Avg} will be further confirmed by following experiments.}
\label{tab:pilot}
\end{table}


\section{Experiments} \label{sec:exp}

\subsection{Experimental Setup} \label{subsec:setup}

\paragraph{Benchmarks} Following \citet{contrastive_nlp_ood}, we choose four datasets corresponding to three tasks as the in-distribution (ID) datasets: SST-2 ~\cite{sst2}  and IMDB~\cite{imdb} for sentiment analysis, TREC-10 \cite{li-roth-2002-learning} for question classification, and 20 Newsgroups \cite{20news} for topic classification. 
Among the four, any pair of datasets coming from different tasks is regarded as an ID-OOD. 
Besides, we use four additional datasets as OOD test data for each ID dataset: WMT-16 \cite{wmt16}, Multi30k \cite{elliott2016multi30k}, RTE \cite{rte}, and SNLI \cite{snli}. 
More details of these datasets can be found in Appendix \ref{app:data1}.

\paragraph{Model Configuration}

We build text classifiers by fine-tuning the RoBERTa-base model~\citep{roberta} (110M parameters) in main experiments. 
Our implementation is based on Hugging Face’s Transformers library \citep{wolf2020transformers}.  
Please refer to Appendix \ref{app:impl} for more details.

\paragraph{Evaluation Protocol}  For OOD detection performance, we report AUROC and FAR95 following \citet{contrastive_nlp_ood}. Higher AUROC and lower FAR95 values indicate better OOD detection performance (specific definitions in Appendix \ref{app:metrics}). 

\begin{table*}[t]
\centering
\resizebox{0.95\textwidth}{!}{
\begin{tabular}{@{}lccccc@{}}
\toprule
\textbf{AUROC$\uparrow$ / FAR95$\downarrow$} & \textbf{SST-2} & \textbf{IMDB} & \textbf{TREC-10} & \textbf{20NG} & \textbf{Avg.} \\ \midrule
\textit{Baselines} & & & & \\
MSP  \citep{maxprob}                 &      88.10 / 70.00          & 96.36 / 22.43  &             94.28 / 24.15     &   87.96 / 49.95   &    91.68 / 41.63    \\
  LOF     \citep{lin2019deep}   &   78.63 / 66.29             &     89.36 / 54.19          &   96.37 / 22.11               &      92.56 / 36.81   &  89.23 / 44.85    \\
  Energy   \citep{liu2020Energy}        &    87.47 / 72.43            &  95.83 / 23.95  & 95.50 / 19.68  &    90.37 / 32.31         &  92.29 /  37.09 \\
MD Baseline    \citep{podolskiy2021revisiting}         &     91.88 / 48.82           &   99.19 / 2.86            &  99.12 / 2.25                 & 96.75 / 15.75     &    96.74 / 17.42      \\
MD + $\mathcal{L}_{\textrm{scl}}$ \citep{contrastive_nlp_ood}                 &  92.16 / 49.04              &    98.86 / 4.08           &  98.59 / 4.96                 &  95.47 / 21.56 & 96.27 / 19.91 \\ 

MD + $\mathcal{L}_{\textrm{margin}}$ \citep{contrastive_nlp_ood}                 &  95.35 / 29.43              &        \textbf{99.93} / \textbf{0.15}      & 99.36 / 1.72                 &  96.51 / 18.51 & 97.79 / 12.45 \\ \midrule
\textit{Ours} & & & & \\

token$=$AVG, layer$=$L12  & 93.14 / 41.24 & 99.67 / 1.09 & 98.29 / 2.88 & 97.07 / 14.07  & 97.04 / 14.82 \\
token$=$CLS, layer$=$ALL & 93.93 / 34.73 & 99.53 / 1.35 & 99.33 / 1.29 & 96.73 / 15.67  &  97.38 / 13.26 \\
token$=$AVG, layer$=$ALL (\textit{Avg-Avg})  & \textbf{97.75} / \textbf{10.67}   & \textbf{ 99.87} / \textbf{0.21}   &\textbf{ 99.66} / \textbf{0.23 } & \textbf{99.70} / \textbf{1.35} &  \textbf{99.25} / \textbf{3.12}

\\ \bottomrule
\end{tabular}}
\caption{The AUROC / FAR95 results of previous OOD detection methods and ours on four benchmarks. $\uparrow$ indicates larger is better and $\downarrow$ indicates lower is better.  For each ID dataset, we report the macro average of AUROC / FAR95 values on all corresponding OOD datasets. All values are percentages averaged over five times with different random seeds, and the best results are highlighted in \textbf{bold}. $\mathcal{L}_{\text{scl}}$ and $\mathcal{L}_{\text{margin}}$ denote the contrastive and margin-based auxiliary  targets proposed by \citet{contrastive_nlp_ood}, respectively. }
\label{tab:main_results}
\end{table*}


\paragraph{Baselines for Comparison} We reimplement a series of existing OOD detection methods for comparison:  MSP \citep{maxprob}, Energy Score \citep{liu2020Energy}, LOF \citep{lin2019deep}, Mahalanobis distance (MD for short) \citep{maha,podolskiy2021revisiting}, and MD combined with contrastive targets \citep{contrastive_nlp_ood}. See Appendix \ref{app:ood_methods} for the introduction and implementation details of these baseline methods.

\subsection{Overall Results} \label{subsec:results}

Table \ref{tab:main_results} gives main results. 
Except the contrastive-based tuning method \citep{contrastive_nlp_ood}, all methods use the same model vanilla fine-tuned on the ID training set. 
Our methods use the Mahalanobis distance to obtain OOD scores, following \citet{podolskiy2021revisiting} (the only difference lies in the embedding space). 
We find that compared to the baseline calculating MD in the last-layer CLS embedding space, both token averaging and layer combination bring improvements on almost all benchmarks. When the two techniques are combined, namely \textit{Avg-Avg} is applied, the performance continues to grow and exceeds the previous state-of-the-art \cite{contrastive_nlp_ood} that needs extra contrastive targets in the fine-tuning stage, by a considerable margin of 9.33\% FAR95 averaged on four benchmarks. 
Further experiments on other PLM backbones also substantiate the enhancement brought by our method, supported by the results in Table~\ref{tab:other_plms}.

\begin{table}[t]
\centering
\resizebox{0.46\textwidth}{!}{
\begin{tabular}{@{}lccccc@{}}
\toprule
\textbf{AUROC $\uparrow$} & \textbf{SST-2} & \textbf{IMDB} & \textbf{TREC-10} & \textbf{20NG} & \textbf{Avg.} \\ \midrule
 ALBERT-base  &  95.43 (+9.62)   &  99.51 (+1.10)  & 98.82 (+0.98)   & 99.55 (+7.02)   &   98.33 (+4.68)  \\ 
 DistillRoBERTa-base &   97.80 (+7.17) & 99.87 (+1.16) & 99.17 (+0.93) & 99.94 (+4.15) & 99.20 (+3.36)  \\ 
BERT-base-uncased      &    95.78 (+2.22)          &  99.51 (+0.45)         &       98.81 (-0.45)         &   99.79 (+1.69)      &  98.45 (+0.95)    \\
RoBERTa-base   &     97.75 (+5.87)  &    99.87 (+0.68)           &   99.66 (+0.54)                &      99.70 (+2.35)    &  99.25 (+2.51)   \\
RoBERTa-large  &  97.49 (+5.13)               &    99.94 (+1.24)           &          99.42 (+0.16)        &    99.97 (+2.88)     & 99.21 (+2.38)      \\ \bottomrule
\end{tabular}}
\caption{The improvements brought by \textit{Avg-Avg} compared to the MD baseline \cite{podolskiy2021revisiting} for different PLMs. AUROC values are reported (the number in the bracket is the improvement).}
\label{tab:other_plms}
\vspace{-0.4cm}
\end{table}

\subsection{Analysis} \label{subsec:analysis}

\paragraph{The Impact of Layer Choice} 

\begin{figure}[t] 
\centering 
\includegraphics[width=0.45\textwidth]{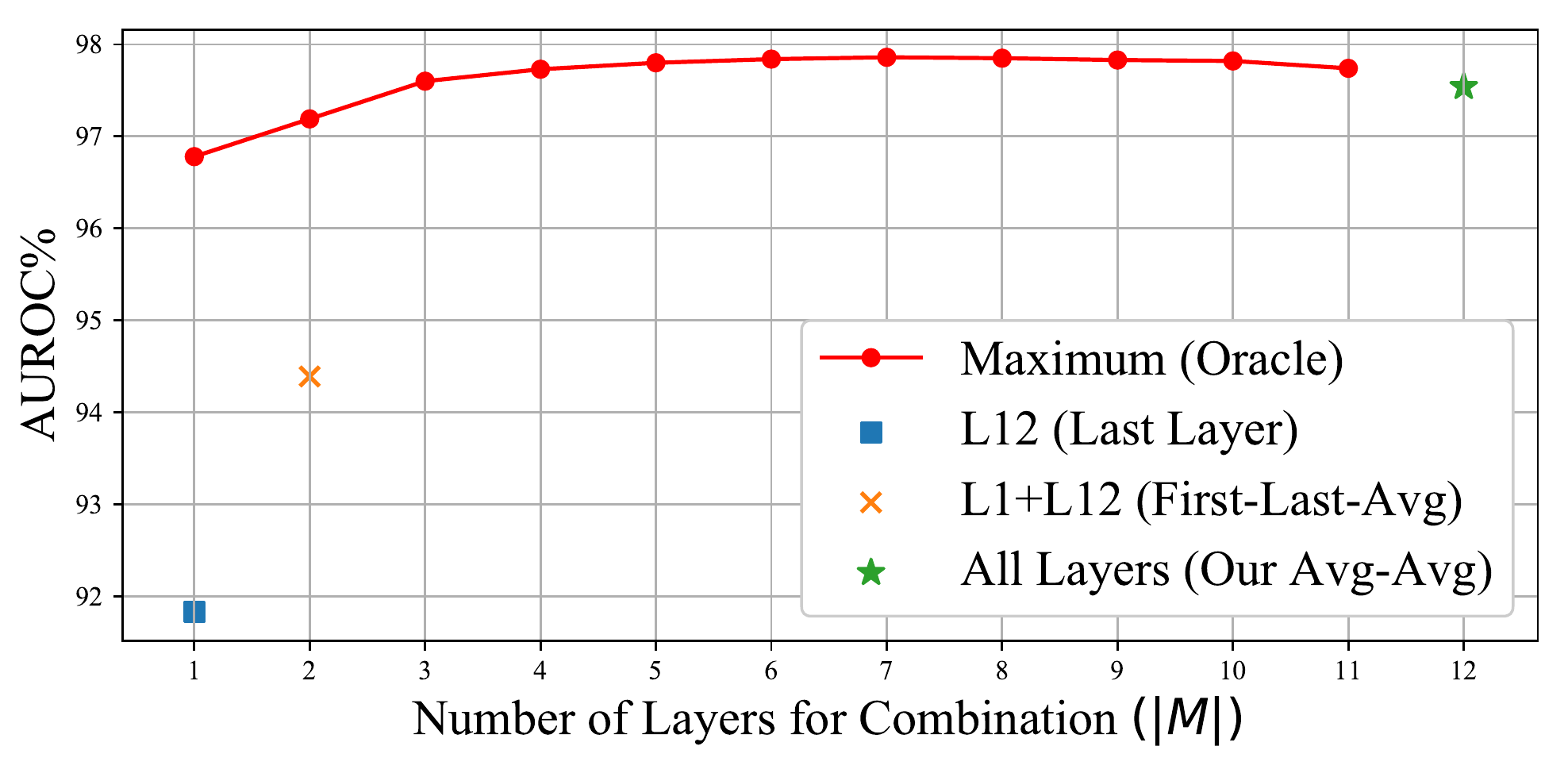}
\caption{Maximum AUROC (averaged over 6 OOD datasets)  values corresponding to sentence embeddings from a RoBERTa model fine-tuned on SST-2 with different numbers of combining layers. The maximum values are results searched on the test data. Token averaging is performed for intra-layer pooling.}
\label{fig:layer_analysis}
\vspace{-0.4cm}
\end{figure}

To verify the rationality of choosing all intermediate layers for inter-layer combination, we show the maximum AUROC values corresponding to different numbers of intermediate layers to derive sentence embeddings in Figure~\ref{fig:layer_analysis}. 
As the number of layers grows, the AUROC metric first increases and then remains relatively stable when more than four layers are chosen. 
Notably, the peak appears when 7 layers are combined, only 0.3\%  higher than our \textit{Avg-Avg}.
Since searching for the best combination of layers is infeasible due to  the unavailability of OOD data, using all layers is a sensible choice.

\paragraph{Probing Analysis}

\begin{table*}[t]
\centering
\resizebox{0.95\textwidth}{!}{
\begin{tabular}{ccccccccccccc}
\toprule
\multirow{2}{*}{\textbf{Intra-Layer}} & \multirow{2}{*}{\textbf{Inter-Layer}} & \multicolumn{2}{c}{\textbf{Surface}} & \multicolumn{3}{c}{\textbf{Syntactic}}                   & \multicolumn{5}{c}{\textbf{Semantic}}                                                   & \multirow{2}{*}{\textbf{Average}} \\ \cmidrule(r){3-4} \cmidrule(r){5-7} \cmidrule(r){8-12}
                                     &                                     & \textbf{SentLen}    & \textbf{WC}    & \textbf{TreeDepth} & \textbf{TopConst} & \textbf{BShift} & \textbf{Tense} & \textbf{SubjNum} & \textbf{ObjNum} & \textbf{SOMO} & \textbf{CoordInv} &                                   \\  \midrule
         CLS      & L12                                 &   48.63             &         10.84    &  26.20              &    49.31          & 74.92            &  83.99       & 76.68             &  72.77          & 57.77       &  62.17        &                  56.33         \\  CLS              & AVG   & 67.74  & 5.55           &  27.86            & 49.53             &  78.88         & 85.43   & 79.75 &  76.32 &  57.67       &      63.45       &                      59.22        \\  
          AVG        & L12   &      61.69         &   12.22     &   30.20         &     51.89         &  79.21       &     85.12     &    78.55         &    76.68      & 59.71      &  62.62      &     59.79                  \\ 
          AVG        & AVG  &    \textbf{91.31}      &  \textbf{17.42}      & \textbf{41.24}      &    \textbf{74.38}   &  \textbf{88.42} & \textbf{88.46}  &  \textbf{84.54}   & \textbf{ 84.54} &  \textbf{63.77}   & \textbf{ 69.14 }      &   \textbf{70.32}   \\ 
                                \bottomrule                     
\end{tabular}}
\caption{Probing task performance for representations corresponding to different pooling strategies. All values are percentages averaged over five RoBERTa models fine-tuned with different random seeds.} 
\label{tab:probing}
\vspace{-0.4cm}
\end{table*}
Given that intermediate layers of PLMs contain a rich hierarchy of linguistic information \cite{jawahar-etal-2019-bert}, a plausible explanation of the performance lift is that \textit{Avg-Avg} leads to an embedding space containing more general linguistic information, where ID and OOD data are more sharply separated. 
To verify this, we evaluate the sentence embeddings produced by the RoBERTa model fine-tuned on SST-2 corresponding to different pooling strategies on the probing tasks proposed by \citet{DBLP:conf/acl/BaroniBLKC18} (details in Appendix \ref{app:data3}). As shown in Table \ref{tab:probing}, our proposed method consistently raises the probing accuracies of surface, syntactic, and semantic level probing tasks, suggesting that we obtain more holistic embeddings by integrating intermediate hidden states.

\begin{table}[t]
\centering
\resizebox{0.48\textwidth}{!}{
\begin{tabular}{ccccc}
\toprule
\multirow{2}{*}{\textbf{\begin{tabular}[c]{@{}c@{}}Dominant\\ Shift\end{tabular}}} & \multirow{2}{*}{\textbf{ID}} & \multirow{2}{*}{\textbf{OOD}} & \multicolumn{2}{c}{\textbf{AUROC}} \\ 
                                                                                     &                              &                               & \textbf{Baseline}  & \textbf{Ours} \\ \midrule
\multirow{2}{*}{Background}                                                          & \multirow{2}{*}{SST-2}       & IMDB                          &            69.86        &    97.57 (+27.70)           \\
                                                                                     &                              & CR           &     75.46   &     82.26 (+6.80)         \\ \midrule
\multirow{2}{*}{Semantic}                                                            & News Top-5                   & News Rest                     &          83.41          &     83.83 (+0.42)              \\  & CLINC                        & CLINC$_{\textrm{OOD}}$                &            97.58           &        97.88 (+0.30)       \\ \bottomrule
\end{tabular}}
\caption{Performance (AUROC) on different kinds of distribution shifts, corresponding to the MD baseline and our proposed \textit{Avg-Avg}. All values are percentages areraged over five different random seeds.}
\label{tab:shift_analysis}
\vspace{-0.3cm}
\end{table}

\paragraph{Detecting Different Kinds of Shifts } OOD texts can be categorized by whether they exhibit a background shift or a semantic shift~\cite{types}.  
In previous main experiments, ID and OOD data come from different tasks and both kinds of shifts exist. 
To explore the source of the performance growth, we conduct ablation experiments by evaluating our method in settings where background or semantic shifts dominate. 
For the semantic shift setting, we use the News Category \cite{misra2018news} and CLINC~\cite{larson-etal-2019-evaluation} datasets (ID and OOD parts share the same background distribution, but belong to different classes); for the background shift setting, we regard SST-2 as ID and IMDB, Customer Reviews (CR for short) \cite{DBLP:conf/kdd/HuL04} as OOD (they all belong to the sentiment analysis task but differ in background features, e.g, the length and style). 
Refer to Appendix \ref{app:data2} for dataset details. 
As shown in Table~\ref{tab:shift_analysis}, our method drastically strengthens the capability of detecting background shifts; in contrast, it only slightly improves detecting semantic shifts, which indicates that the performance gain mainly comes from the task-agnostic general linguistic information in the holistic embeddings obtained by our pooling technique, in line with the probing analysis.

\subsection{Comparison with Universal Sentence Embedding Approaches}

Here we further show the advantage of our method \textit{Avg-Avg}
over two representative universal sentence embedding approaches, SentenceBERT (SBERT)~\citep{reimers-gurevych-2019-sentence} and SimCSE~\citep{gao2021simcse} on OOD detection. 
For SBERT, we test the model trained on NLI (natural language inference) data (last-layer mean pooling is adopted as recommended in the original work); for SimCSE, we test the unsupervised model and the supervised model trained on NLI data (last-layer CLS pooling is adopted). 
The backbone model is RoBERTa-base in all methods.
We also fine-tune the pre-trained models on the ID data and use the default pooling ways and our \textit{Avg-Avg} to obtain embeddings from fine-tuned models for thorough comparison on OOD detection.
As results in Table~\ref{tab:rebuttal_results}, when \textit{Avg-Avg} is applied, it brings consistent improvements and beats both pre-trained and fine-tuned sentence embedding models using the default pooling way.
These results corroborate 
the advantage of \textit{Avg-Avg} as a specialized embedding method for OOD detection.
\begin{table}[t]
\centering
\resizebox{0.46\textwidth}{!}{
\begin{tabular}{@{}lrrrrr@{}}
\toprule
\textbf{Method} & \textbf{SST-2} & \textbf{IMDB} & \textbf{TREC} & \textbf{20NG} & \textbf{Avg.} \\ \midrule
SBERT  & 21.02 & 2.40 & 19.03 & 3.63 & 11.52 \\
SBERT$_{\textrm{ft}}$  & 35.02 & 5.65  &  0.78 & 12.13 & 13.40 \\
unsup-SimCSE & 42.02 &  9.51 & 62.68 & \textbf{0.00} & 28.55 \\
unsup-SimCSE$_{\textrm{ft}}$ & 44.43 & 2.30 & 2.69 & 19.56 & 17.25 \\
sup-SimCSE & 32.37 & 3.16 & 35.42 & 0.03 & 17.75 \\
sup-SimCSE$_{\textrm{ft}}$  & 38.91 & 0.59 & 1.30  & 15.14 & 13.99 \\
vanilla+\textit{last-cls}           &     48.82           &  2.86            &  2.25     & 15.75     &     17.42     \\ \midrule
SBERT$_{\textrm{ft}}$+\textit{Avg-Avg}  & 9.70  & 0.33   & 0.11 & 1.67 &  3.03
 \\
unsup-SimCSE$_{\textrm{ft}}$+\textit{Avg-Avg} & 11.36 & 0.42 & 0.40 & 1.13 &  3.33
\\  
sup-SimCSE$_{\textrm{ft}}$+\textit{Avg-Avg} & \textbf{7.68} & \textbf{0.18}  & \textbf{0.14}  & 0.77 & \textbf{2.19}  
\\ 
vanilla+\textit{Avg-Avg}  & 10.67  & 0.21   & 0.23 & 1.35 & 3.12 \\ 
\bottomrule
\end{tabular}}
\caption{The OOD detection performance (FAR95) of different embedding approaches (lower FAR95 values indicate better detection performance). The ``ft'' subscript denotes that the embedding model is fine-tuned on the in-distribution data for classification.}
\label{tab:rebuttal_results}
\vspace{-0.3cm}
\end{table}


\subsection{Embedding Visualization}

\begin{figure}[t]
    \centering
    \begin{subfigure}[The default last-layer CLS vectors.]{ \label{subfig:tsne_default}
        \includegraphics[width=0.45\textwidth]{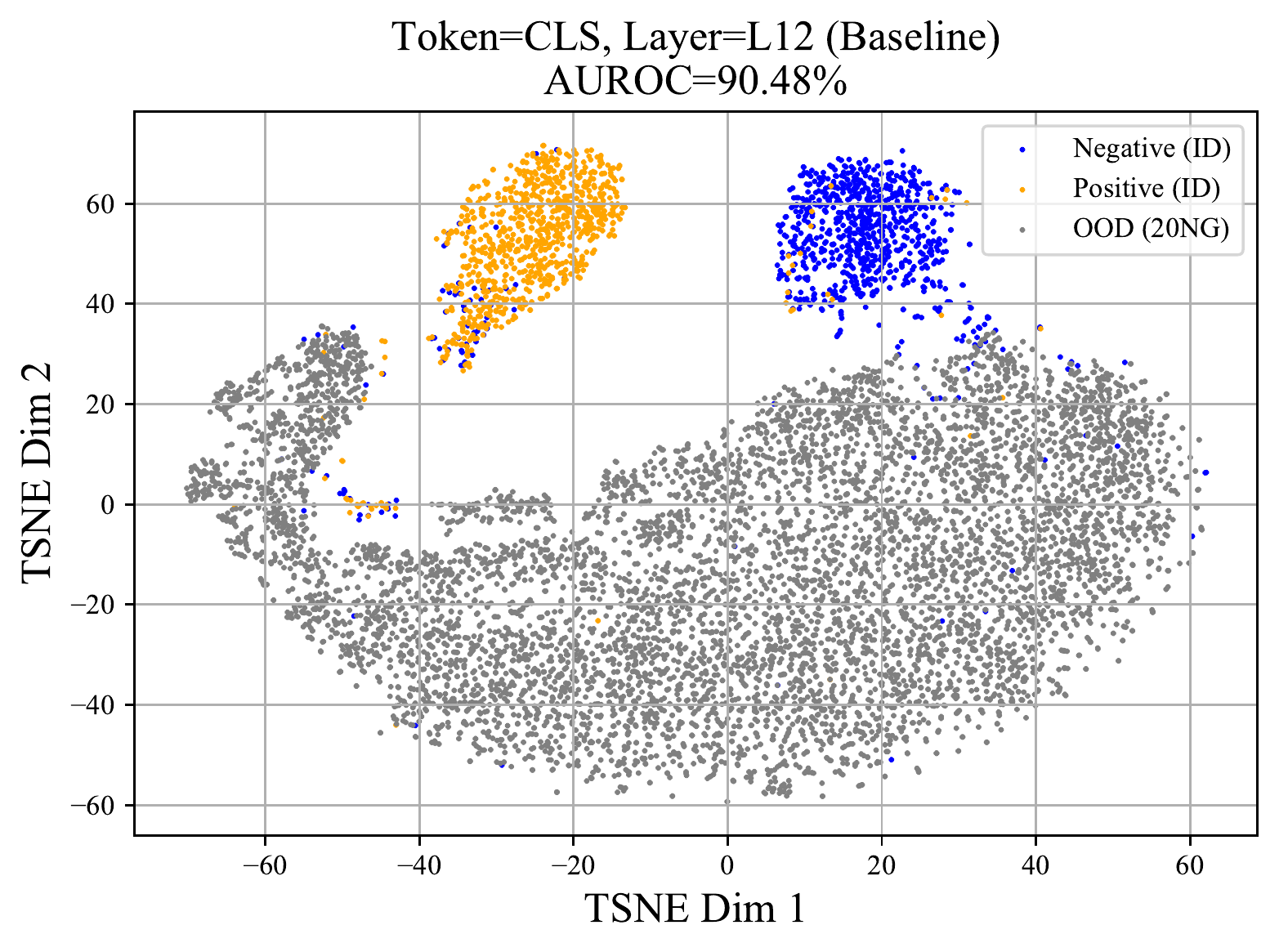}
    }
    \end{subfigure} 
    
    \begin{subfigure}[Our \textit{Avg-Avg} embeddings.]{ 
    \label{subfig:tsne_ours}
        \includegraphics[width=0.45\textwidth]{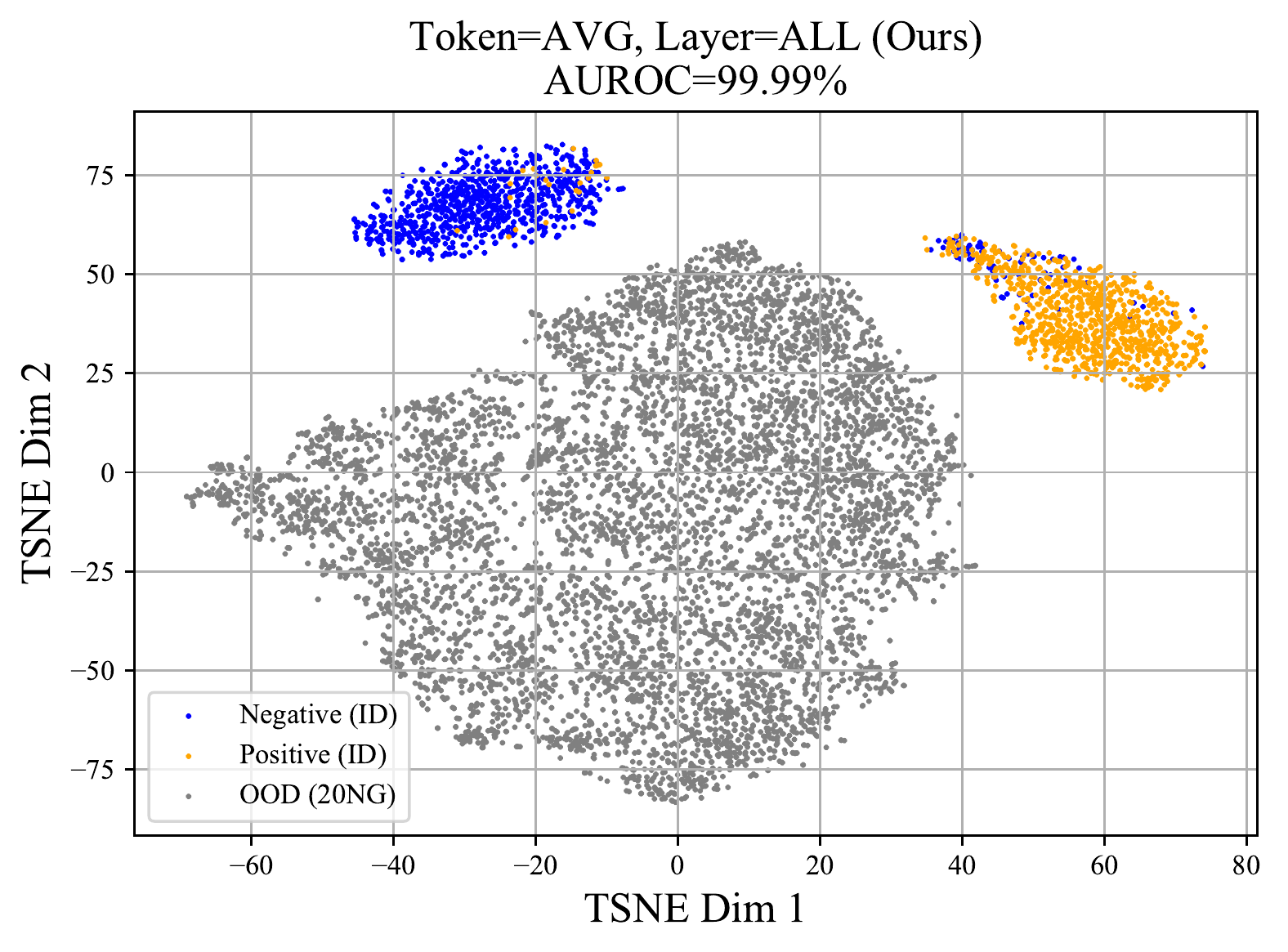}
    }
    \end{subfigure}
    
    \caption{Visualization of the representations obtained for \textcolor{orange}{positive}, \textcolor{blue}{negative} instances in SST-2 and \textcolor{gray}{OOD} ones (20 Newsgroups).  }
    \label{fig:tsne}
     
\end{figure}

To demonstrate the influence of  the studied pooling strategies on the embedding space, we fine-tune the RoBERTa-base model on SST-2 and visualize instance embeddings corresponding to different pooling strategies from the SST-2 test set (ID) and an OOD test set (20 Newsgroups) using t-SNE \cite{van2008visualizing}. As plotted in Figure~\ref{fig:tsne}, in the representation space produced by \textit{Avg-Avg} (Figure~\ref{subfig:tsne_ours}) where is almost no overlap between ID and OOD instances, ID and OOD samples are more sharply separated than in the space of default last-layer CLS embeddings (Figure~\ref{subfig:tsne_default}).
This further supports our claim that \textit{Avg-Avg} is better suited for OOD detection.

\section{Related Works} \label{sec:related_works}

\subsection{Textual OOD Detection}

OOD  detection  aims  to  detect abnormalities  that  come  from  a  different  distribution  from the  training  set \cite{maxprob}. 
Compared with the widely studied OOD image detection problem \cite{odin,maha,liu2020Energy,fssd,fort2021exploring,yang2021oodsurvey}, textual OOD detection remains under-explored. 
\citet{hendrycks2020pretrained} first showed that pretrained Transformers improved OOD detection using the maximum softmax probability \cite{maxprob}.
Afterward, \citet{podolskiy2021revisiting} used the Mahalanobis distance approach \citep{maha} for PLM-based OOD detection and achieved superior performance. Following this, \citet{contrastive_nlp_ood} further raised the performance by utilizing contrastive auxiliary targets in the fine-tuning stage.

\subsection{Unsupervised Sentence Embedding}

Unsupervised sentence embedding is a well-established area~\citep{kiros2015skip,pagliardini2017unsupervised,BERT-FLOW,reimers-gurevych-2019-sentence,gao2021simcse}. 
Relevant to our work, \citet{su2021whitening} and \citet{huang-etal-2021-whiteningbert-easy} proposed to obtain better sentence embeddings via averaging token representations, layer combination, and a whitening operation. 
It is noteworthy that these embedding approaches are mainly studied for sentence matching and retrieval tasks. 
As far as we know, we are the first to study novel embedding ways to replace the default last-layer CLS pooling  for boosting textual OOD detection.

\section{Conclusion} \label{sec:conclusion}

In this work, we focus on how to derive sentence embeddings suitable for OOD detection from fine-tuned PLMs. 
Specifically, we introduce token averaging and layer combination to derive more holistic representations and substantially improve the capability of PLMs to detect OOD inputs. 
Moreover, our analysis shows that our approach helps preserve general linguistic information and benefits detecting background shifts. 
Overall, our work points out a new perspective that textual OOD detection can be enhanced by obtaining high-quality sentence embeddings, and we hope to extend this idea to training-time methods in future work.

\section*{Limitations}

The contemporary solution \textit{Avg-Avg} is primarily motivated by empirical observations and its effectiveness is confirmed by extensive experiments on different PLMs and benchmarks. 
Currently, its superiority lacks strict theoretical justifications and there is still a small performance gap between our method and the ideal upper bound as shown in Figure \ref{fig:layer_analysis}.
In future work, we plan to explore theory-guaranteed embedding approaches to further boost the OOD detection ability of PLMs.


\section*{Ethical Considerations}
Our work presents an efficient embedding method to enhance the OOD detection ability of NLP models. 
We believe that our proposal
will help reduce security risks resulting from OOD inputs to NLP models deployed in the open-world environment. 
In addition, all experiments in this
work are conducted on open datasets and our code is publicly available. 
While we do not expect any direct negative consequences
to the work, we hope to continue exploring more efficient and robust sentence embedding approaches for textual OOD detection  in future work.

\section*{Acknowledgement}
We sincerely thank all the anonymous reviewers
for their constructive comments and suggestions.
This work is supported by Natural Science Foundation of China (NSFC) No. 62176002.
Xu Sun is the corresponding author of this paper.

\bibliography{anthology,custom}
\bibliographystyle{acl_natbib}

\appendix

\section{Dataset Statistics and Introduction} \label{app:data}

\subsection{Datasets Used in Main Experiments} \label{app:data1}

The statistics of in-distribution (ID) and out-of-distribution (OOD) textual datasets in main experiments (Section \ref{subsec:setup} and \ref{subsec:results}), including the number of classes,  the dataset size, and the average length of samples, are given in Table \ref{tab:id_stats} and \ref{tab:ood_stats}, respectively.
Here is a brief introduction to these datasets: Multi30k \cite{elliott2016multi30k} and WMT16 \cite{wmt16} are parts of the English side data of English-German machine translation datasets; RTE \cite{rte} and SNLI \cite{snli} are the concatenations of the precise and respective hypotheses from NLI datasets.

\begin{table}[t]
\centering
\resizebox{0.45\textwidth}{!}{
\begin{tabular}{@{}lrrrrr@{}}
\toprule
\textbf{Dataset} & \textbf{\#Classes} & \textbf{\#Train} & \textbf{\#Dev} & \textbf{\#Test} & \textbf{L} \\ \midrule
SST-2            &     2             &                6920      &        872         &    1821              &        19    \\
IMDB             &        2          &               23000       &        2000         &  25000                &      230      \\
TREC-10       &     6               &         4907        &      545    &      500           &   10        \\
20 Newsgroups          &      20              &          10182       &      1132          &   7532             &     289    \\ \bottomrule
\end{tabular}}
\caption{Statistics of in-distribution text datasets. \textbf{L} denotes the average length of samples.}
\label{tab:id_stats}
\end{table}
\begin{table}[t] \small
\centering
\begin{tabular}{@{}lrr@{}}
\toprule
\textbf{Dataset} & \textbf{\#Test} & \textbf{L} \\ \midrule
Multi30k         &     1014                  & 13 \\
WMT16            &   2000                & 22           \\
RTE              &      3000            & 48            \\
SNLI             &      2000            &  21          \\ \bottomrule
\end{tabular}
\caption{Statistics of out-of-distribution text datasets. \textbf{L} denotes the average length of samples.}
\label{tab:ood_stats}
\end{table}

\subsection{Datasets Used In the Distribution Shift Analysis } \label{app:data2}

\citet{types} categorized the distribution shifts in natural language data into two main types: background shifts and semantic shifts.
We follow their division and study OOD detection performance  under the setting where either kind of shift dominates in Section \ref{subsec:analysis}. 
The statistics of extra datasets used in the distribution shift analysis are given in Table \ref{tab:shift_data_stats}.
Here is a brief introduction to these datasets.

\paragraph{Background Shift Setting.} Background shifts refer to the shift of background features (e.g., formality) that do not depend on the label.
We consider domain shifts in sentiment classification datasets. 
SST-2 contains short movie reviews by the audience, while IMDB contains longer and more professional movie reviews. 
Customer Reviews \cite{DBLP:conf/kdd/HuL04} contains reviews for different kinds of commercial products on the web, representing a domain shift from SST-2. 
So the IMDB and Customer Reviews test data can be regarded as OOD samples for the model fine-tuned on SST-2.  

\paragraph{Semantic Shift Setting.} 
In this setting, OOD data are from the same task as ID data and share similar background characteristics, but belong to classes unseen during training.
We use the News Category \cite{misra2018news} and the CLINC \cite{clinc} datasets to create two ID/OOD pairs under the setting. 
Following \citet{types}, we use the data from the five most frequent classes of the News Category as ID (News Top-5) and the data from the remaining 36 classes as OOD (News Rest). 
In the CLINC dataset for intent classification,  there is a 150-class ID subset and an OOD test set CLINC$_{\textrm{OOD}}$ composed of utterances belonging to actions not supported by existing ID intents.

\begin{table}[t]
\centering
\resizebox{0.48\textwidth}{!}{
\begin{tabular}{@{}lrrrrr@{}}
\toprule
\textbf{Dataset} & \textbf{\# Classes} & \textbf{\# Train} & \textbf{\# Dev} & \textbf{\# Test} & \textbf{L} \\ \midrule
Customer Reviews (OOD) &  2 &  - & - & 1000 & 20  \\
News Top-5 (ID)            &     5             &     68859   &       8617          &     8684          & 30          \\
News Rest (OOD) & 36  & -  & - & 11402 & 29 \\
CLINC (ID) & 150 & 15000 & 3000 & 4500 & 8 \\
CLINC$_{\textrm{OOD}}$ (OOD) & - & - & - & 1000   & 9 
 \\ \bottomrule
\end{tabular}}
\caption{Statistics of extra datasets introduced for the distribution shift analysis. \textbf{L} denotes the average length of each sample.}
\label{tab:shift_data_stats}
\end{table}

\subsection{Probing Benchmarks} \label{app:data3}

To probe the linguistic information contained in sentence embeddings, we use the probing tasks proposed by \citet{DBLP:conf/acl/BaroniBLKC18}, which are grouped into three categories. For surface information, we use \textit{SentLen} (sentence length) and  \textit{WC} (the presence of words); for syntactic information, we use \textit{BShift} (sensitivity to word order), \textit{TreeDepth} (the depth of the syntactic tree), and \textit{TopConst} (the sequence of top-level constituents); for semantic information, we use \textit{Tense} (tense), \textit{SubjNum} and \textit{ObjNum} (the subject/direct object number in the main clause), \textit{SOMO} (the sensitivity to random replacement of a noun/verb), and \textit{CoordInv} (the random swapping of coordinated clausal conjuncts). Each probing dataset contains 100k training samples, 10k validation samples, and 10k test samples. 
We use the SentEval toolkit \cite{conneau-kiela-2018-senteval} along with the recommended hyperparameter space to search for the
best probing classifier according to the validation accuracy and report test accuracies.

\section{Details of Pretrained Language Model Fine-tuning} \label{app:impl}

\subsection{Vanilla Fine-tuning} We use the RoBERTa-base pretrained model \citep{roberta} as the backbone to build text classifiers by fine-tuning it on the ID training data. We use a batch size of 16 and fine-tune the model for 5 epochs. The model is optimized with the Adam \citep{adam} optimizer using a learning rate of 2e-5. We evaluate the model on the ID development set after every epoch and choose the best checkpoint as the final model. The setting is the same for other pretrained Transformers studied in the paper (RoBERTa-large, BERT-base-uncased, DistilRoBERTa-base, and ALBERT-base). DistilRoBERTa  \cite{Sanh2019DistilBERTAD} is a light distilled RoBERTa and ALBERT \cite{lan2019albert} is a lite BERT with factorized embedding parameterization and cross-layer parameter sharing.

\subsection{Contrastive Auxiliary Targets}

\citet{contrastive_nlp_ood} introduced two alternatives of contrastive loss to boost textual OOD detection, i.e., the supervised contrastive loss and the margin-based contrastive loss. For a classification task with $C$ classes, given a batch of training examples $\{x_i,y_i\}_{i=1}^M$, where $x_i$ is the input and $y_i$ is the label, the supervised contrastive loss term $L_{
\textrm{scl}}$ and the final optimization target $\mathcal{L}$ can be formulated as:

\begin{equation}
\resizebox{0.42\textwidth}{!}{$
\begin{aligned}
 & \mathcal{L}_{\textrm{scl}} =\sum_{i=1}^{M} \frac{-1}{M|P(i)|} \sum_{p \in P(i)} \log \frac{e^{\boldsymbol{z}_{i}^{\top} \boldsymbol{z}_{p} / \tau}}{\sum_{a \in A(i)} e^{\boldsymbol{z}_{i}^{\top} \boldsymbol{z}_{a} / \tau}}, 
\\
 & \mathcal{L} = \mathcal{L}_{\textrm{ce}} +\mathcal{L}_{\textrm{scl}}
\end{aligned}$}
\end{equation}

where $A(i) = \{1, ...,M\} \setminus \{i\}$ is the set of all anchor instances, $P(i) = \{ p \in A(i): y_i = y_p \} $ is the set of anchor instances from the same class  as $i$, $\tau$ is a temperature hyper-parameter, $\boldsymbol{z}$ is the L2-normalized CLS embedding before the softmax layer, $\mathcal{L}_{\text {ce}}$ is the cross entropy loss, and $\lambda$ is a positive coefficient.
Following \citet{contrastive_nlp_ood}, we use $\tau=0.3$ and $\lambda=2$.

The margin-based loss term $\mathcal{L}_{\text {margin }}$ and the final optimization target $\mathcal{L}$ can be formulated as:

\begin{equation}
\resizebox{0.42\textwidth}{!}{
$
\begin{aligned}
&\mathcal{L}_{\textrm{pos }}=\sum_{i=1}^{M} \frac{1}{|P(i)|} \sum_{p \in P(i)}\left\|\boldsymbol{h}_{i}-\boldsymbol{h}_{p}\right\|^{2}, \\
&\mathcal{L}_{\textrm{neg }}=\sum_{i=1}^{M} \frac{1}{|N(i)|} \sum_{n \in N(i)}\left(\xi-\left\|\boldsymbol{h}_{i}-\boldsymbol{h}_{n}\right\|^{2}\right)_{+}, \\
&\mathcal{L}_{\textrm{margin }}=\frac{1}{d M}\left(\mathcal{L}_{\textrm{pos }}+\mathcal{L}_{\textrm{neg }}\right), \\
&\xi=\max _{i=1}^{M} \max _{p \in P(i)}\left\|\boldsymbol{h}_{i}-\boldsymbol{h}_{p}\right\|^{2}, \\
&\mathcal{L} = \mathcal{L}_{\textrm{ce}} + \lambda \mathcal{L}_{\textrm{margin}}.
\end{aligned}$}
\end{equation}

Here $N(i)=\left\{n \in A(i): y_{i} \neq y_{n}\right\}$ is the set of anchor instances from other classes than $y_{i}, \boldsymbol{h} \in \mathbb{R}^{d}$ is the unnormalized CLS embedding before the softmax layer, $\xi$ is the margin, $d$ is the number of dimensions of $\boldsymbol{h}$, and $\lambda$ is a positive coefficient. We use $\lambda=2$ following \citet{contrastive_nlp_ood}.

Except for the loss term, we use the same hyper-parameters for these two tuning methods as vanilla tuning. Table~\ref{tab:acc} gives test accuracies on four ID datasets for the RoBERTa models tuned with vanilla cross-entropy loss ($\mathcal{L}_{\textrm{ce}}$), supervised contrastive loss ($\mathcal{L}_{\textrm{ce}} + \mathcal{L}_{\textrm{scl}}$), and margin-based contrastive loss ($\mathcal{L}_{\textrm{ce}}+\mathcal{L}_{\textrm{margin}}$), where are not significant differences.

\subsection{Hardware Requirements}

All the experiments (fine-tuning and inference) in this paper are conducted on a single NVIDIA TITAN RTX GPU, except that the fine-tuning of the RoBERTa-large model needs 4 TITAN RTX GPUs.

\section{Definition of Evaluation Metrics for OOD Detection} \label{app:metrics}

For an input instance $\mathbf{x}$, the output of an OOD detector is the confidence score $S(\mathbf{x})$ (a higher confidence score). A higher confidence score indicates that the detector tends to regard $\mathbf{x}$ as a normal ID sample. In real applications, system users need to choose a threshold $\gamma$ and treat the OOD detection module as a binary classifier:

\begin{table}[t]
\centering
\resizebox{0.45\textwidth}{!}{
\begin{tabular}{@{}lrrrr@{}}
\toprule
\textbf{Loss} & \textbf{SST-2} & \textbf{IMDB} & \textbf{TREC} & \textbf{20NG}  \\ \midrule
      $\mathcal{L}_{\text{ce}}$        &      93.96        &  94.56              &         95.88      &   84.52            \\
       $\mathcal{L}_{\text{ce}} + \mathcal{L}_{\text{scl}}$       &   94.23             &   94.53            &         96.36      &  84.65            \\
      $\mathcal{L}_{\text{ce}} + \mathcal{L}_{\text{margin}}$          &   93.69             &          94.21     &        95.76       &   84.63       \\ \bottomrule
\end{tabular}}
\caption{Test accuracies on four ID datasets for RoBERTa-base models tuned with three different fine-tuning strategies. All values are percentages averaged over five times with different random seeds.}
\label{tab:acc}
\end{table}

\begin{equation}
\centering
G(\mathbf{x})= \begin{cases}\textrm{in,} & \textrm{if } S(\mathbf{x}) \geq \gamma \\ \textrm{out,} & \textrm{if } S(\mathbf{x})<\gamma\end{cases}
\end{equation}

Following previous works \cite{maxprob,contrastive_nlp_ood}, we use the following two threshold-free metrics for evaluation:

\textbf{AUROC} is short for the area under the receiver operating curve. It plots the true positive rate (TPR) against the false positive rate (FPR) and can be interpreted as the probability that the model ranks a random positive(ID)  example more highly than a random negative (OOD) example. A higher AUROC indicates better OOD detection performance.

\textbf{FAR95} is the probability for a negative example (OOD) to be mistakenly classified as positive (ID) when the TPR is 95\%. A lower value indicates better detection performance.

\section{OOD Detection Baselines} \label{app:ood_methods}

\subsection{MSP}

MSP \cite{maxprob}  is a classical baseline using the maximum softmax probability in the prediction outputs of the classifier as the confidence score, i.e., $S(\mathbf{x}) = \max_{y \in \Upsilon} p_{y}(\mathbf{x})$. 

\subsection{Energy Score}

\citet{liu2020Energy} proposed using free energy as a scoring function for OOD detection. For a classification problem with $C$ classes, a multi-class classifier $f(\mathbf{x}): \mathcal{X} \rightarrow \mathbb{R}^{C}$ can be interpreted from an energy-based perspective by viewing the logit output $f_{y_i}(\mathbf{x})$ corresponding to class $y_i$ as an energy function $E(\mathbf{x},y_i)=-f_{y_i}(\mathbf{x})$. The free energy function $E(\mathbf{x})$ for an input $\mathbf{x}$ is $E(\mathbf{x})= \sum_{i=1}^{C} e^{f_{y_{i}}(\mathbf{x})}$, and $S(\mathbf{x}) = -E(\mathbf{x})$.

\subsection{LOF}

\citet{lin2019deep} proposed identifying unknown user intents by feeding feature vectors to the density-based novelty detection algorithm, local outlier factor (LOF) \cite{breunig2000lof}. We use the last-layer CLS vectors produced by the fine-tuned RoBERTa models as the input and train a LOF model following the implementation details of \citet{lin2019deep} on the ID training set and use the local density output as $S(\mathbf{x})$. 

\begin{table*}[t]
\centering
\resizebox{0.9\textwidth}{!}{
\begin{tabular}{@{}lccccc@{}}
\toprule
\textbf{AUROC$\uparrow$ / FAR95$\downarrow$}& \textbf{SST-2} & \textbf{IMDB} & \textbf{TREC-10} & \textbf{20NG} & \textbf{Avg} \\ \midrule
Maha Baseline   &     91.88 / 48.82           &   99.19 / 2.86            &  99.12 / 2.25                 & 96.75 / 15.75     &    96.74 / 17.42          \\
SE, token=CLS  &        94.68 / 26.09        &   \textbf{99.94} / \textbf{0.05}              &     \textbf{99.75} / \textbf{0.19}              & 99.47 / 2.62         &  98.45 / 7.24      \\
SE, token=AVG  &      97.19 / 13.58          &   99.84 / 0.33             & 99.50 / 0.38                 &   \textbf{99.82} / \textbf{0.83}      &    99.09 / 3.78     \\
\textit{Avg-Avg} (Ours)      & \textbf{97.75} / \textbf{10.67}   & 99.87 / 0.21  & 99.66 / 0.23 & 99.70 / 1.35 &  \textbf{99.25} / \textbf{3.12} \\ \bottomrule
\end{tabular}}
\caption{Comparison between score ensemble (SE) and  \textit{Avg-Avg}.The setting of backbone models and ID/OOD benchmarks is the same as that in Table \ref{tab:main_results}. }
\label{tab:score_ensemble}
\end{table*}

\subsection{Mahalanobis Distance}

Mahalanobis distance score \cite{maha} is a representative distance-based OOD detection algorithm, which uses the sample distance to the nearest ID class in the embedding space as the OOD uncertainty measure. For a given feature extractor $\psi$, the Mahalanobis distance score is defined as: $
S(\mathbf{x})=\max_{c \in \Upsilon}  -\left(\psi(\mathbf{x})-\mu_{c}\right)^{T} \Sigma^{-1}\left(\psi(\mathbf{x})-\mu_{c}\right)$, where $\psi(\mathbf{x})$ is the embedding vector of the input $\mathbf{x}$, $\mu_{c}$ is the class centroid for a class $c$, and $\Sigma$ is the covariance matrix. 
The estimation of $\mu_{c}$ and $\Sigma$ is defined as follows:

\begin{equation}
\small
\begin{gathered}
\mu_{c}=\frac{1}{N_{c}} \sum_{\mathbf{x} \in \mathcal{D}_{\textrm{in}}^{c}} \psi(\mathbf{x}), \\
\Sigma=\frac{1}{N} \sum_{c \in \Upsilon} \sum_{\mathbf{x} \in \mathcal{D}_{i\textrm{in}}^{c}}\left(\psi(\mathbf{x})-\mu_{c}\right)\left(\psi(\mathbf{x})-\mu_{c}\right)^{T},
\end{gathered}
\end{equation}

where $\mathcal{D}_{\textrm{in}}^{c}=\left\{\mathbf{x} \mid(\mathbf{x}, y) \in \mathcal{D}_{\textrm{in}}, y=c\right\}$ denotes the training samples belonging to the class $c$, $N$ is the size of the training set, and $N_{c}$ is the number of training instances belonging to class $c$.

\section{Comparison with Score Ensemble}

Apart from the layer combination technique studied in the paper, there is another way to utilize intermediate representations for OOD detection: estimating the sample distance score in the embedding space of each intermediate layer and taking their sum as the final OOD score.  
For the Mahalanobis distance score, the final ensemble score $S(\mathbf{x})$ is defined as:

\begin{small}
\begin{equation}
\begin{gathered}
S^{\ell}(\mathbf{x})=\max_{c \in \Upsilon}  -\left(\psi^{\ell}(\mathbf{x})-\mu_{c}^{\ell}\right)^{T} \Sigma^{-1}_{\ell}\left(\psi^{\ell}(\mathbf{x})-\mu_{c}^{\ell}\right), \\
S(\mathbf{x}) = \sum_{\ell} \alpha_{\ell}S^{\ell}(\mathbf{x}),
\end{gathered}
\end{equation}
\end{small}

where $\psi^{\ell}(\mathbf{x})$ denotes the output features at the $\ell$th-layer of neural networks, and $\mu^{\ell}$ and $\Sigma_{\ell}$ are the class mean and the covariance matrix, correspondingly. The layer-wise weighting hyperparameter is $\alpha_{\ell}$. In the original work \cite{maha}, $\alpha_{\ell}$ is tuned  on a small validation set containing both ID and OOD for each OOD dataset, which is impractical in the setting of unsupervised OOD detection followed by recent works (OOD data is not available). Following \citet{godin}, we use uniform weighting, i.e., $S(\mathbf{x}) = \sum_{\ell} S^{\ell}(\mathbf{x})$, in the baselines for comparison.

We compare the performance of  SE (score ensemble) and \textit{Avg-Avg} and show the results in Table \ref{tab:score_ensemble}. We observe that SE also brings consistent improvements over the baseline using only last-layer CLS vectors. Without token averaging, SE slightly surpasses \textit{Avg-Avg} on IMDB and TREC-10, but underperforms \textit{Avg-Avg} significantly on SST-2 and 20NG; when token averaging is performed, SE only beats \textit{Avg-Avg} on 20NG but underperforms on other three benchmarks, especially remarkably on SST-2. 
In view of the average performance on the four benchmarks, we can get \textit{Avg-Avg} > SE (AVG) > SE (CLS). 
Considering that the class mean $\mu^{\ell}$  and the inverse of covariance matrix $\Sigma^{-1}_{\ell}$ need to be estimated and stored for each layer in SE, \textit{Avg-Avg} is also more convenient for deployment. So compared with SE, \textit{Avg-Avg} enjoys both simplicity and advantages in performance.

\end{document}